\documentclass[letterpaper, 10 pt, conference]{ieeeconf}  

\IEEEoverridecommandlockouts                              
\usepackage[utf8]{inputenc}
\usepackage[T1]{fontenc}

\usepackage{amsmath} 
\usepackage{amssymb}  
\usepackage{hyperref}
\usepackage{cleveref}
\usepackage[detect-all,per-mode=symbol]{siunitx}
\usepackage{booktabs}
\usepackage{multirow}
\usepackage{pgfplots}
\usepackage[protrusion=true,expansion=true]{microtype} 
\usepackage[keeplastbox]{flushend}
\usepgfplotslibrary{groupplots}
\usetikzlibrary{positioning}
\usepackage[noend]{algcompatible}
\usepackage{algorithm}
\newcommand{\argmax}{\operatornamewithlimits{arg\,max}}
\usepackage{subcaption}
\usepackage{float}

\usepackage[style=ieee, maxcitenames=2, mincitenames=1,natbib=true,dashed=false]{biblatex}
\pgfplotsset{compat=newest}

\usepackage{soul,color}
\usepackage{booktabs}

\title{\LARGE \bf
Reinforcement Learning for Autonomous Driving with \\ Latent State Inference and Spatial-Temporal Relationships
}

\author{Xiaobai Ma$^{1,2}$, Jiachen Li$^{1,3}$, Mykel J. Kochenderfer$^{2}$, David Isele$^{1}$, and Kikuo Fujimura$^{1}$
\thanks{*This work was supported by Honda Research Institute USA, Inc. (HRI).}
\thanks{$^{1}$Honda Research Institute US, \texttt{\{disele, kfujimura\}@honda-ri.com}}
\thanks{$^{2}$Stanford University, \texttt{ \{maxiaoba, mykel\}@stanford.edu}
}%
\thanks{$^{3}$University of California, Berkeley, \texttt{jiachen\_li@berkeley.edu}}%
}

\begin{document}

\maketitle
\thispagestyle{empty}
\pagestyle{empty}

\begin{abstract}
Deep reinforcement learning (DRL) provides a promising way for learning navigation in complex autonomous driving scenarios. However, identifying the subtle cues that can indicate drastically different outcomes remains an open problem with designing autonomous systems that operate in human environments.
In this work, we show that explicitly inferring the latent state and encoding spatial-temporal relationships in a reinforcement learning framework can help address this difficulty. 
We encode prior knowledge on the latent states of other drivers through a framework that combines the reinforcement learner with a supervised learner. In addition, we model the influence passing between different vehicles through graph neural networks (GNNs). 
The proposed framework significantly improves performance in the context of navigating T-intersections compared with state-of-the-art baseline approaches.
\end{abstract}

\section{Introduction}
Deep reinforcement learning (DRL) has shown promising performance on various control and decision making tasks including robotics~\cite{gu2017deep}, games~\cite{lample2017playing}, and autonomous driving~\cite{isele2018navigating,mohseni2019interaction}. Compared with traditional rule-based~\cite{idm,mobil} or optimization-based approaches~\cite{cesari2017scenario,lima2015clothoid} to autonomous driving, DRL methods have the potential for better scalability and generalization in complex scenarios that require observing subtle changes in behavior and executing complex interactions with other agents \cite{yang2018cm3,iteKQ,saxena2020driving}. DRL is better suited to these problems due to the high representational capability of neural networks ~\cite{codevilla2018end,sadigh2016planning}, better generalization to model mismatch ~\cite{lin2020comparison}, and the ability to efficiently model approximate solutions to intractable combinatorial relationships \cite{xu2018powerful}. However, DRL is known to require large amounts of data and is still prone to unexpected behavior on out-of-sample scenarios. 

One approach to address anomalous behavior is to decompose the system into engineered sub-modules. Each sub-module can be treated as a more controlled and more rigorously verified learning sub-problem existing as either stand-alone units \cite{redmon2016you} or hybrid objectives \cite{liang2019multi}. These sub components have behavior that is often easier to analyze, which greatly facilitates debugging complex systems. 

Our work studies whether an auxiliary learning problem \cite{jaderberg2016reinforcement} for inferring the latent state of traffic participants would extract useful information that an unassisted learning system would otherwise miss. We consider several frameworks for incorporating auxiliary information into a decision making process, and we show that, with an improper framework, extra information can act as a distractor for learning even if it is useful information for the target task. Also, we show that the auxiliary inference can have mutually beneficial coupling with relational modeling. Specifically we show that graph nets improve latent inference in the auxiliary task, while the latent inference improves the quality of the learned policy.
\begin{figure}[!t]
    \centering
    \includegraphics[width=0.9\columnwidth]{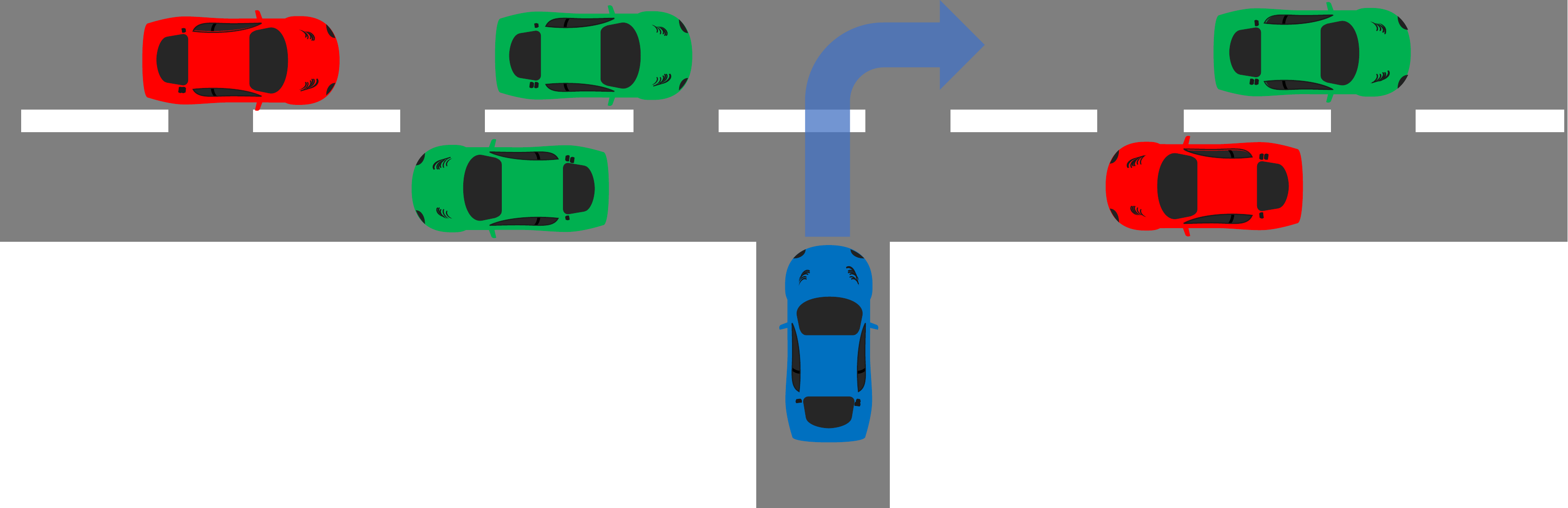}
    \caption{\small \textbf{T-intersection scenario}: The ego vehicle (blue) is trying to make a right turn and merge into the further lane without colliding the conservative (green) and aggressive (red) vehicles.\normalsize}
    \vspace{-0.7cm}
    \label{fig:t_intersection}
\end{figure}
We focus on the problem of autonomous driving through a T-intersection as illustrated in \cref{fig:t_intersection}.
The human drivers’ behaviors are driven by their latent characteristics, which are usually not directly observable. When entering the intersection, it is important to know whether the on-coming vehicles are willing to yield to the ego vehicle. Successfully inferring the latent states of other drivers can help the ego driver make better decisions. While most reinforcement learning methods implicitly learn the latent states through returns~\cite{matiisen2018deep}, we show that learning this explicitly will improve the efficiency and interpretability of the learning.

As drivers are influenced by their surrounding vehicles, different traffic participants may directly or indirectly impact other drivers. For example, in highway driving, the ego driver is more directly influenced by neighboring vehicles, while vehicles farther away influence the ego driver indirectly through chains of influence that propagate to the neighbors. Such influence-passing forms a graph representation of the traffic scenario, where each node represents a traffic participant and each edge represents a direct influence. With the recent development of graph neural networks (GNN)~\cite{graphsage,gat}, we are able to model such relational information contained in graphs efficiently.

By combining the latent inference with relational modelling, we propose a reinforcement learning framework that explicitly learns to infer the latent states of surrounding drivers using graph neural networks. The proposed framework significantly improves the performance of the autonomous driving policy in the free T-intersection, while also providing labels of latent states and influence passing structure that help interpretability. 
The contributions of this paper are as follows:
\begin{itemize}
    \item We propose a reinforcement learning framework augmented with an additional supervised learning process that uses the additional prior knowledge at training time.
    \item We show how explicitly learning to infer the latent states of surrounding drivers can accelerate reinforcement learning in autonomous navigation tasks.
    \item We investigate the applications of graph representations and graph neural networks on driver latent inference as well as vehicle control.
    \item Our proposed framework shows significant improvement in the free T-intersection navigation task compared with state-of-the-art baselines.
\end{itemize}

\section{Related Work}

There is a rich literature on applying deep reinforcement learning methods to address various challenges in autonomous navigation tasks.
\citet{chen2019model} investigate the application of different model-free reinforcement learning approaches on urban driving scenarios.
\citet{mohseni2019interaction} use multi-agent reinforcement learning to model the interactions between different drivers.
\citet{isele2018navigating} explore the use of deep Q learning \cite{mnih2013playing} on the intersection scenario with an emphasis on occluded vehicles.

The latent states of other traffic participants including intentions and driving styles are often recognized as important factors for decision making in the autonomous driving context~\cite{bai2015intention,song2016intention,dong2017intention}. 
Existing work has explored the combination of a POMDP solver with various latent inference methods including the un-parameterized belief tracker~\cite{bai2015intention} and the hidden Markov model (HMM)~\cite{song2016intention}. While a POMDP solver is able to consider the uncertainty in the intention inference, it is computationally expensive. 
Deep learning methods are also applied to the latent state inference as a standalone task~\cite{kumar2013learning,xing2020ensemble}.
\citet{morton2017simultaneous} investigates the simultaneous learning of the latent space and the vehicle control through imitation learning.
In our work, both the latent inference and the vehicle control are learned under a reinforcement learning framework to handle complex observations and scenarios.


GNNs are a type of deep learning model that is directly applied to graph structures. They naturally incorporate relational inductive bias into learning-based models \cite{battaglia2018relational}.
GNNs have been used to model interactions between multiple interactive agents in different application domains including on-road vehicles \cite{gao2020vectornet,cao2021SpecTGNN}, pedestrians \cite{huang2019stgat,li2020social,mohamed2020social}, sports players \cite{li2020Evolvegraph}, and robotics~\cite{wang2018nervenet}.
These works mainly focus on behavior modeling, trajectory prediction, as well as joint control, while we are leveraging the graph structure to infer the latent state for decision making tasks.

\begin{figure*}[!tbp]
	\centering
	\includegraphics[width=0.8\textwidth]{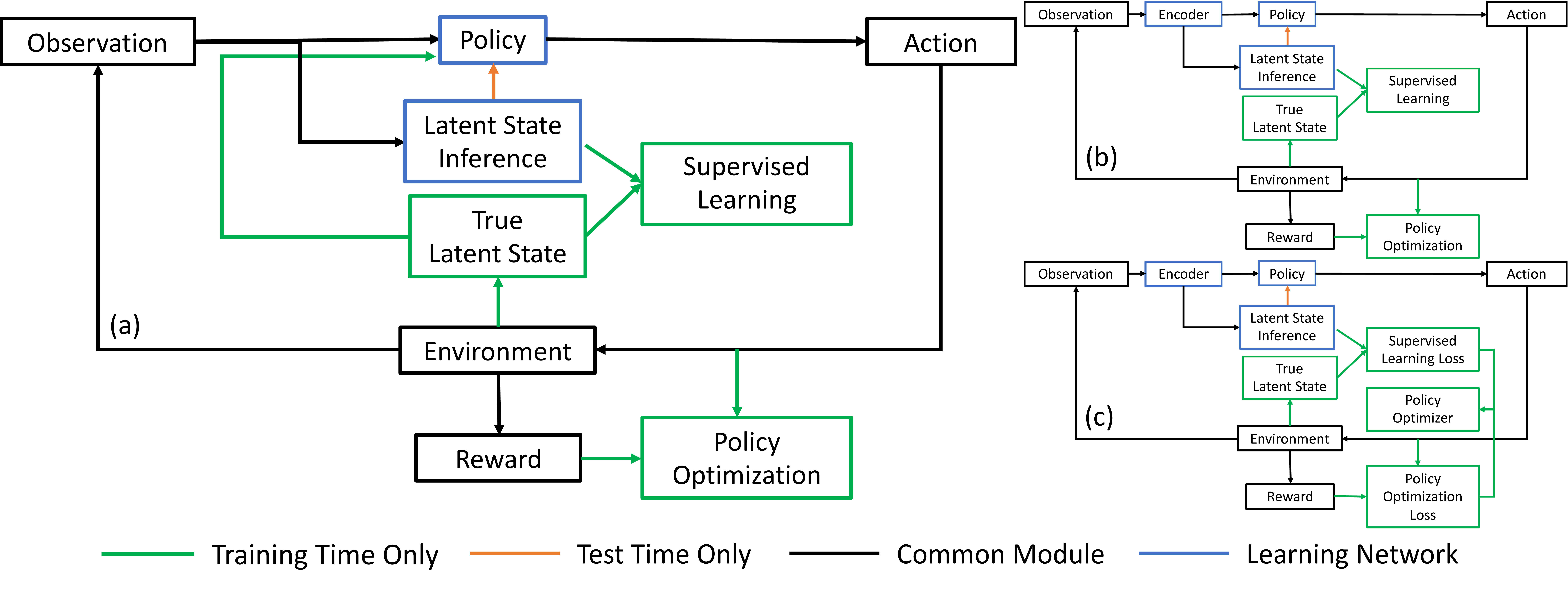}
	\vspace{-0.3cm}
	\caption{\small System architecture for policy optimization with auxiliary latent inference. Common modules in black are present during both training and test time. (a) Separated inference (proposed); (b) Shared inference; (c) Coupled inference.\normalsize
	}\label{fig:structure}
	\vspace{-0.7cm}
\end{figure*}
\section{Preliminaries}
\subsection{Partially Observable Markov Decision Process}
A Markov decision process (MDP)~\cite{dmu} describes the sequential decision making process of an agent interacting with an environment. It is specified by the tuple $(S,A,T,R,\gamma, \rho_0)$, where $S$ is the state space; $A$ is the action space; $T$: $S \times A \times S \rightarrow \mathbb{R}$ is the transition model; $R: S \times A \rightarrow \mathbb{R}$ is the reward model; $\gamma \in [0,1]$ is the discount factor; and $\rho_0: S \rightarrow \mathbb{R}$ is the initial state distribution. When the state is not directly observable, a partially observable Markov decision process (POMDP) can be used by augmenting the MDP with an additional observation function $\Omega:S \rightarrow O$. $\Omega$ maps a state $s\in S$ to an observation $o\in O$, where $O$ is the observation space. The objective in a POMDP is to find a policy $\pi:O\rightarrow A$ that maximizes expected return:
\begin{equation}
    \pi^* = \argmax_\pi \mathbb{E}_{s_0,a_0,o_0,\ldots}\sum_{t=0}^{\infty}\gamma^tR(s_t,a_t)
\end{equation}
where $s_0\sim \rho_0(s_0)$, $a_t\sim \pi(a_t\mid o_{1:t})$, $o_t\sim \Omega(o_t\mid s_t)$, and $s_{t+1}\sim T(s_{t+1}\mid s_t,a_t)$.

\subsection{Policy Optimization}
\label{sec:po}
Policy optimization methods work by directly optimizing the policy parameters through policy gradients~\cite{reinforce,trpo,ppo}. The REINFORCE method provides an unbiased gradient estimator using the objective 
$
    L^{PG}(\theta)=\hat{\mathbb{E}}[\log\pi_\theta(a,s)\hat{A}]
$,
where $\hat{\mathbb{E}}$ is the empirical average over the experiences collected using the most up-to-date policy parameters $\theta'$, and $\hat{A}$ is the estimated advantage.
For a POMDP, the state is usually replaced by the observation and the hidden state of the policy. 
In PPO~\cite{ppo}, a clipped surrogate objective is maximized:
\begin{equation}
\label{eq:ppo}
    L^{PPO}(\theta)=\hat{\mathbb{E}}[\min(r(\theta)\hat{A},\text{clip}(r(\theta),1-\epsilon,1+\epsilon)\hat{A})]
\end{equation}
where $r(\theta)=\frac{\pi_\theta(a\mid s)}{\pi_{\theta'}(a\mid s)}$ is the probability ratio with $\theta'$ being the old policy parameter used to collect the experience. $\epsilon$ is the clipping range. This clipping constrains the update step size which usually provides more stable performance than REINFORCE.

\section{Methods}
\subsection{Auxiliary Latent Inference}
\label{sec:aux}
We use the latent state inference as an auxiliary task to improve the primary reinforcement learning task. We use PPO~\cite{ppo} as the core RL algorithm for its stability and computational efficiency.

Consider the driving scenario containing the ego vehicle and $n$ surrounding vehicles, where the ego vehicle is controlled by the learning agent and the surrounding vehicles are controlled by $n$ human drivers. Let $x_t$ denote the physical state of all vehicles at timestep $t$, we model the action distribution of the $i$th human driver as $P(a^i_t\mid x_t,z^i_t)$, where $z^i_t$ represents the latent state of the driver.

The goal of the auxiliary latent inference is to learn $P(z^i_t\mid o_{1:t})$, where $o_{1:t}$ is the ego agent's historical observation up to time $t$. We assume that the true latent states of the surrounding drivers are known at training time and unknown at the test time. Thus, a latent inference network could be trained through supervised learning.
By utilizing the information provided by the latent state labels, the auxiliary intention inference provides additional learning signals compared to a traditional reinforcement learner.

Inferring the latent state instead of directly predicting the future trajectories of the surrounding drivers provides several advantages: First, a latent state is often simpler and more efficient to learn as well as  encode into the control policy; Second, in most cases, the latent state provides the same or even more information than the predicted trajectories. In the T-intersection scenario illustrated in \cref{fig:t_intersection}, the predicted trajectories of the conservative and aggressive vehicles could be similar at the moment before the ego vehicle approaches the intersection. However, their latent states could be inferred by observing their interaction histories with other vehicles. In this case, the latent state provides the key information for the ego vehicle to predict the future behaviors of surrounding drivers where the predicted trajectories could not.

In this work, we treat the policy and the latent inference as two separated modules which are trained simultaneously. 
The latent inference network learns the mapping from the historical observations to the latent distribution, i.e. $P_\phi(z^i_t\mid o_{1:t})$ where $\phi$ contains the parameters of the latent inference network trained to maximize the log-likelihood objective:
\begin{equation}
    L(\phi)=\mathbb{E}_{z^i_t,o_{1:t}\sim D}[\log P_\phi(z^i_t\mid o_{1:t})]
\end{equation}
where the true latent state $z^i_t$ and the observation history $o_{1:t}$ are randomly sampled from a replay buffer storing exploration experiences.
Training the latent inference module simultaneously with the policy module helps improve the data efficiency as well as minimizes the distribution shift between the training experiences of the two modules.

The policy takes input of both the observation history and the latent state, i.e. $\pi_\theta(a\mid o_{1:t},z^{1:n}_t)$, where $\theta$ contains the policy network parameters trained by the augmented policy optimization objective:
\begin{align}
\label{eq:ppo2}
    L(\theta)=\hat{\mathbb{E}}[\min(&\frac{\pi_\theta(a\mid  o_{1:t},z^{1:n}_t)}{\pi_{\theta'}(a\mid  o_{1:t},z^{1:n}_t)}\hat{A}, \nonumber\\
    &\text{clip}(\frac{\pi_\theta(a\mid o_{1:t},z^{1:n}_t)}{\pi_{\theta'}(a\mid o_{1:t},z^{1:n}_t)},1-\epsilon,1+\epsilon)\hat{A})]
\end{align}

At training time, we feed the ground truth latent state to the policy for exploration as well as policy optimization. 
At test time, we use the latent state inferred by the latent inference network. This brings several benefits: First, feeding the ground truth latent state at exploration time helps the policy find the trajectory leading to the task goal. This is especially important when the task is difficult and the reward is sparse. Second, by using a separate network for each task, we minimize the mutual influence of the gradients from different tasks because such mutual influence could be harmful as shown in our experiments. Third, by modularizing the two learning parts, our framework allows flexible choices for network structures in different modules. 
The proposed framework is illustrated in \cref{fig:structure}(a).

We also investigate two variants of the architecture shown in \cref{fig:structure}(b) and (c).
In \cref{fig:structure}(b), an encoder is shared between the policy network and latent inference network, but the two tasks are trained with separate losses and optimizers. While in \cref{fig:structure}(c), besides the shared encoder, the losses from two tasks are also coupled by a weighted sum and the entire network is trained with the policy optimization optimizer.
These variants are employed as baselines in the experiments.

\begin{figure*}	
	\centering
	\includegraphics[width=0.8\textwidth]{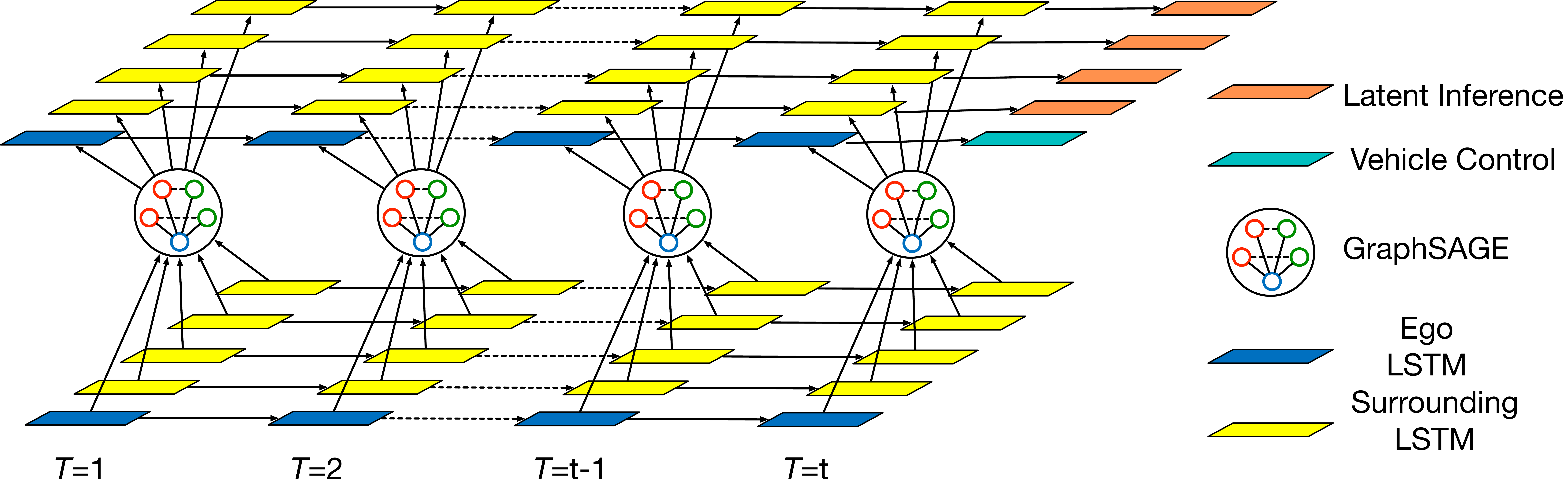}
	\vspace{-0.2cm}
	\caption{\small The diagram of the proposed STGSage architecture.\label{fig:graph} \normalsize}
	\vspace{-0.7cm}
\end{figure*}

\subsection{Graph Representation}
\label{sec:graph}
A driver's behavior on the road is heavily influenced by its relations to other traffic participants. The shared dependence of different traffic participants can be formalized as a graph with each car represented as a node. 
For the vehicles driving on the main road of the T intersection, the drivers are only directly influenced by the closest vehicles in its lane as well as the ego vehicle which is trying to merge. 
The ego vehicle are considered to be influenced by all vehicles to make the optimal long-term plan. 
Based on this intuition, we represent the T-intersection scenario at time $t$ as a graph $\mathcal{G}_t = (\mathcal{V}_t,\mathcal{E}_t)$, where the node set $\mathcal{V}_t$ contains the nodes for all the vehicles in the scene, and the edge set $\mathcal{E}_t$ contains all the direct influence between vehicles. 

We adopt a three layer network architecture similar as STGAT~\cite{huang2019stgat} to process both the spatial relational information in $\mathcal{G}_t$ and the temporal information in $o_{1:t}$.
The proposed model architecture, named as STGSage, are illustrated in \cref{fig:graph}. 
At timestep $t$, the observation on the $i$th vehicle, $o^i_t$, as well as its observation history, $o^i_{1:t-1}$ represented by the LSTM hidden state $h^i_t$, are fed into the corresponding bottom level LSTM. The LSTM parameters are shared for all the vehicles except the ego vehicle.
The output of the LSTM is used as the initial embedding for the corresponding vehicle node in $\mathcal{G}_t$. The node embeddings are then updated by several layers of GraphSage~\cite{graphsage} convolution. \Cref{subsec:gnn_compare} gives a detailed comparison and discussion of the advantage of GraphSage instead of GAT as used in STGAT.
The updated embedding from the graph convolutions is then fed to the corresponding top level LSTM with the same parameter sharing scheme.

STGSage could be used as the policy network as well as the latent inference network. When used as the policy network, the action distribution could be calculated from the output of the top level ego LSTM network. When served as the latent inference network, the output of the top level LSTM of each surrounding vehicle is used to calculate the corresponding latent state. \Cref{subsec: network_compare} discusses the performance of STGSage in both tasks.

\section{Problem Formulation}
\label{sec:prob_form}
The uncontrolled T-intersection scenario is framed as a POMDP.
As illustrated in~\cref{fig:t_intersection}, multiple vehicles are driving in a two-lane roadway where the upper lane goes to right and the lower lane goes to left. The ego vehicle in blue is trying to merge into the upper lane from a third vertical lane. The goal is to maintain the smoothness of the traffic while making a successful right turn. 

\subsubsection{State}
The physical state of the scenario is defined as $x = [x^0, x^1, x^2, \ldots, x^n]$, where $x^0\in \mathbb{R}^4$ is the state of the ego vehicle containing its position and velocity, $x^i\in \mathbb{R}^4$ is the state of the $i$th vehicle. The number of surrounding vehicles varies as vehicles enter into or exit from the scenario.
The latent state of the surrounding drivers is represented as $z = [z^1, z^2, \ldots, z^n]$, where $z^i\in \{\textsc{Conservative},\textsc{Aggressive}\}$ indicates the driving style of the $i$th driver. 
The joint state is then represented by 
$$s = [x^0,(x^1,z^1), (x^2,z^2), \ldots, (x^n,z^n)]$$

\subsubsection{Observation}
The physical states of all surrounding vehicles are observable to the ego agent,  but the latent state are not. We have 
$o = [x^0+\epsilon^0, x^1+\epsilon^1, x^2+\epsilon^2, \ldots, x^n+\epsilon^n]$,
where $\epsilon^i$ is a small observation noise added to the observation on the physical state of the $i$th vehicle sampled from a zero-mean Gaussian.

\subsubsection{Action}
The action $a\in\{\SI{0}{\m\per\second},\SI{0.5}{\m\per\second},\SI{3}{\m\per\second}\}$ controls the target speed of the ego vehicle's low level controller.

\subsubsection{Transition}
The simulation proceeds with an interval of \SI{0.1}{\second} between consecutive time steps.
The surrounding vehicles are controlled by the IDM model~\cite{idm} with a desired speed of \SI{3}{\meter\per\second}. The actual acceleration of the surrounding vehicles are also influenced by a Gaussian noise with a standard deviation of $0.1$. 

In practice, human agents can exhibit a broad variety of subtle cues indicating their future behavior \cite{li2020Evolvegraph}, and modeling that behavior accurately in simulation remains an open problem. To ensure these subtle differences exist, but also give us tight control over the experiment, we model differences in behavior with overlapping distributions of the front gap size.
Upon noticing the ego vehicle, the surrounding driver would shrink the gap with its leading vehicle to prevent the ego vehicle merging into its lane. 
Conservative drivers vary their desired front gap uniformly between $0.5$ to $0.8$ of the original gap, where the aggressive driver varies between $0.4$ to $0.7$. 
The conservative drivers would yield to the ego driver if they intercept laterally or the ego vehicle is approaching the lane center with a speed larger than \SI{0.5}{\meter\per\second}, while the aggressive driver would ignore the ego vehicle. A driver is sampled to be \textsc{conservative} or \textsc{aggressive} from a binary distribution with $P(\textsc{conservative})=0.5$ at the beginning of the episode.
The difference in the desired front gap provides the ego driver with evidence on which to predict, whether the surrounding vehicles will yield or not before the actual merging. As the divergence between the two desired gap distributions is small, it is very difficult for a reinforcement learner to infer the intentions of the driver implicitly. 
 
The ego vehicle is controlled by a longitudinal PD controller which follows the right-turn trajectory and tracks the desired speed set by the POMDP actions. It also has a safety checker that performs an emergency break if the ego vehicle is too close to other vehicles. 
The episode ends when the ego vehicle makes a full right-turn, a collision happens, or the maximum horizon of 200 time steps is reached.

\begin{figure*}	
	\begin{picture}(200.0, 100.0)
    	\put(5,  0){\includegraphics[width=.95\columnwidth]{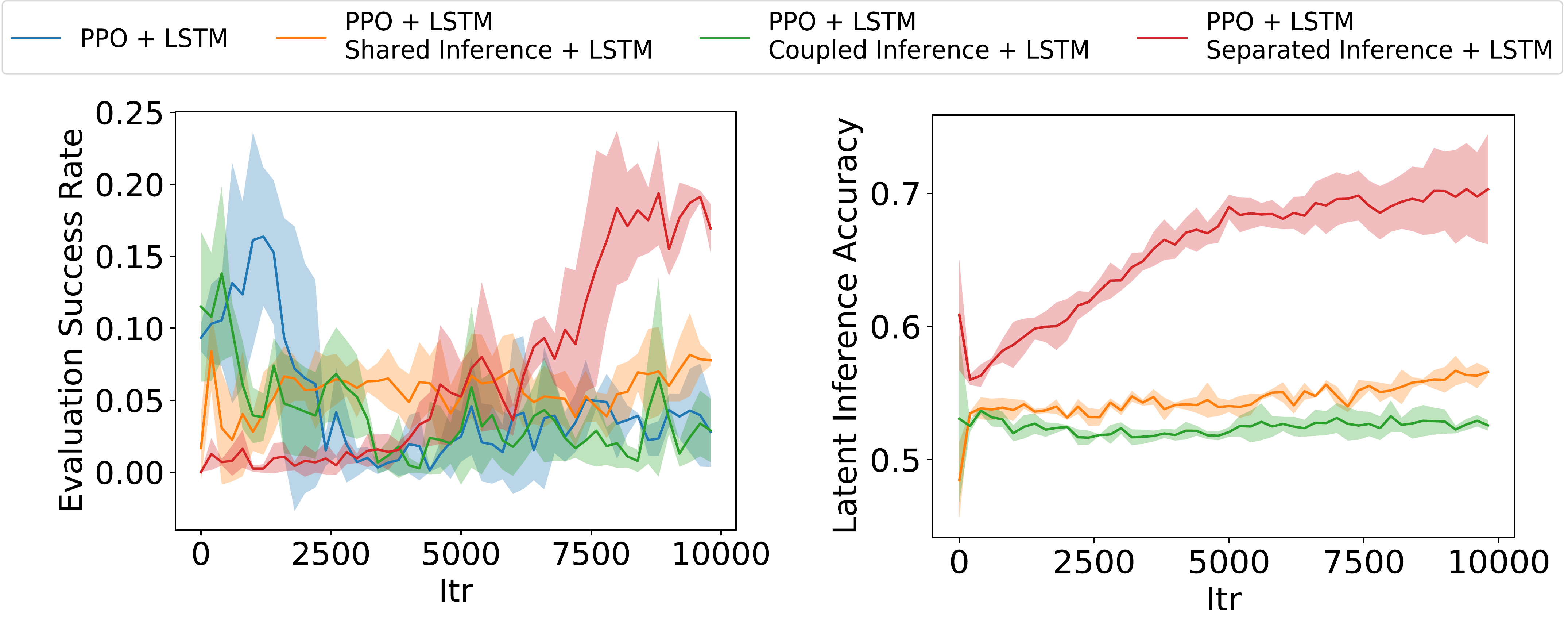} }
    	\put(255,  0){\includegraphics[width=1.01\columnwidth]{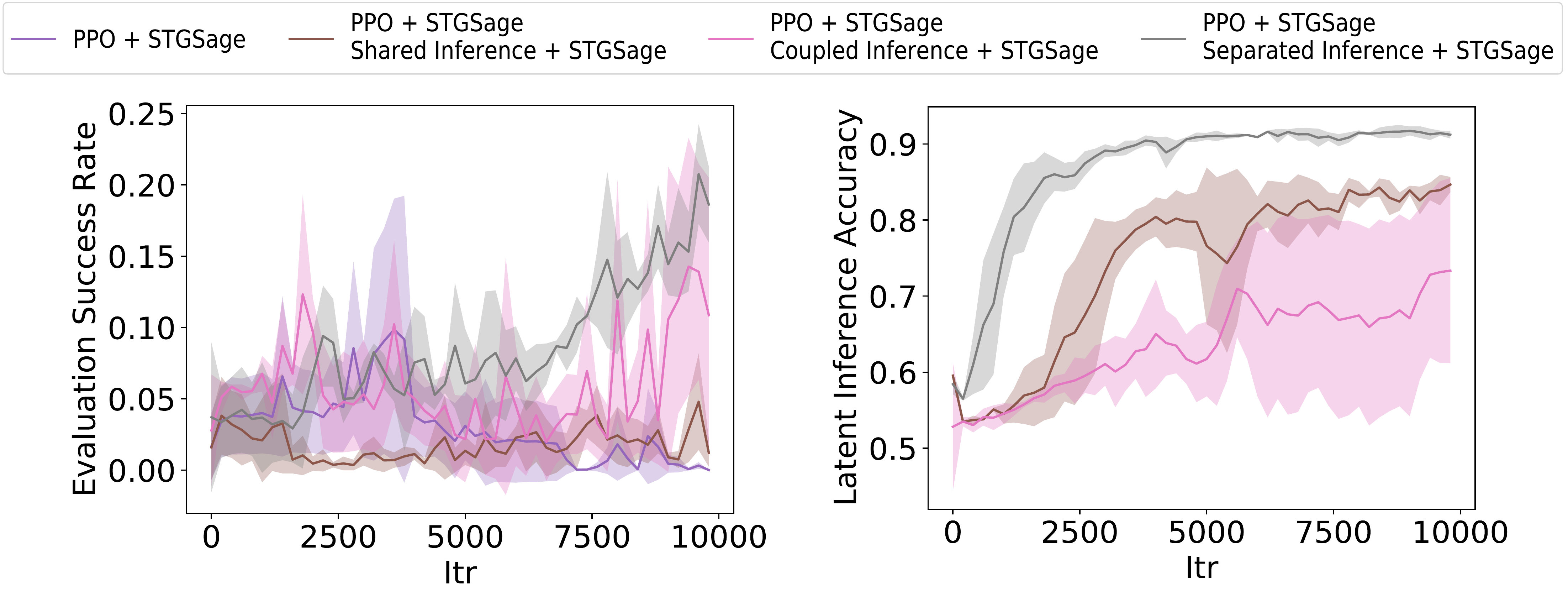} }
    	\put(15,5){(a)}
        \put(130,5){(b)}
        \put(270,5){(c)}
        \put(390,5){(d)}
	\end{picture}
	\vspace{-0.3cm}
	\caption{\small \textbf{Latent Inference architecture comparison}: Influence of the auxiliary latent state inference and difference inference architectures on the performance of PPO and latent inference.
	(a-b): Architecture comparison using LSTM;
	(c-d): Architecture comparison using STGSage; (a), (c): Test time success rate. The separated inference improves the success rate by 140\% on LSTM and 70\% on STGSage; (b), (d): Latent state inference accuracy. The separated inference improves the inference accuracy by 30\% on LSTM and 10\% on STGSage.\normalsize}
	\label{fig:structure_compare}
	\vspace{-0.5cm}
\end{figure*}

\subsubsection{Reward}
The reward function is designed to encourage the learning agent to control the vehicle passing the intersection as quick as possible without collisions, which is calculated by:
\small $
    R(s,a) = r_\text{goal}\mathbf{1}(s \in S_{goal})+r_\text{fail}\mathbf{1}(s \in S_\text{fail}) + r_\text{speed}(s)
$
\normalsize,
where $r_\text{goal}=2$ and $S_\text{goal}$ is the set of goal states where the ego vehicle successfully makes a full right-turn; $r_\text{fail}=-2$ and $S_\text{fail}$ is the set of failure states where there is a collision; $r_\text{speed}(s)=0.01\frac{v_{ego}}{\SI{3}{\meter\per\second}}$ is a small reward on the speed of the ego vehicle.

\section{Experiments}

We run multiple experiments comparing different combinations of training architectures as shown in \cref{fig:structure} as well as network structures.
Each setting is repeated with 3 trials using different random seeds, and each trial is trained for 10k epochs with 1k environment steps per epoch. 
We use a learning rate of \num{e-4} for the policy optimizer and \num{e-3} for the value baseline and the supervised learner in the shared and the separated inference structure. 
In the coupled inference structure, we set the weight of the supervised learning loss to 0.1.
The input to the LSTM network is a concatenation of the observations on all vehicles and is ordered such that the features of the same vehicle always occupy the same position in the observation vector throughout the episode. The input to the STGSage network is first divided by vehicles and then fed to the corresponding low level LSTMs. 
We use one hidden layer for all the LSTMs and three convolutional layers for all the GNNs.
All the policy action generation heads and the intention inference heads use a linear layer.
We use 48 hidden units in the LSTM for PPO as well as PPO with the shared/coupled inference. We use 28 hidden units for the separated inference structure. For STGSage, we use a node dimension of 24 in GraphSage and 24 hidden units in the sub-LSTMs for PPO and the shared/coupled inference. The node dimension and the number of hidden units are reduced to 18 for the separated inference. The number of parameters in all networks tested are close to \num{2e4}.
Our implementation is based on the \textsc{rlkit}.\footnote{https://github.com/vitchyr/rlkit} 

\subsection{Latent State Inference and Training Architecture}

We test the influence of the auxiliary latent inference on the performance of PPO with our proposed separated inference structure.
We also compare it with the two variants shown in \cref{fig:structure}.
As shown in \cref{fig:structure_compare} (a) and (c), for both LSTM and STGSage, the proposed separated inference structure significantly improves the reinforcement learning performance compared with all baselines. 
The additional latent state labels provide more information of the environment than a traditional reinforcement learning algorithm could get from the rollouts. 
When this additional information is indeed important for the reinforcement learning task, the performance gain is expected.
However, we see less improvement on the RL performance by using the shared or coupled inference structure. If we compare the inference performance shown in \cref{fig:structure_compare}(b) and \cref{fig:structure_compare}(d), the separated structure also achieves a much better accuracy than the two baselines. 
This result indicates that the effect of the feature shaping from the auxiliary task is not always helping, which is likely due to the distraction caused by the multiple loss sources so that the gradient estimates on both tasks are biased. Such distraction is minimized in the separated structure.

\begin{figure}[!tbp]
	\centering
	\includegraphics[width=\columnwidth]{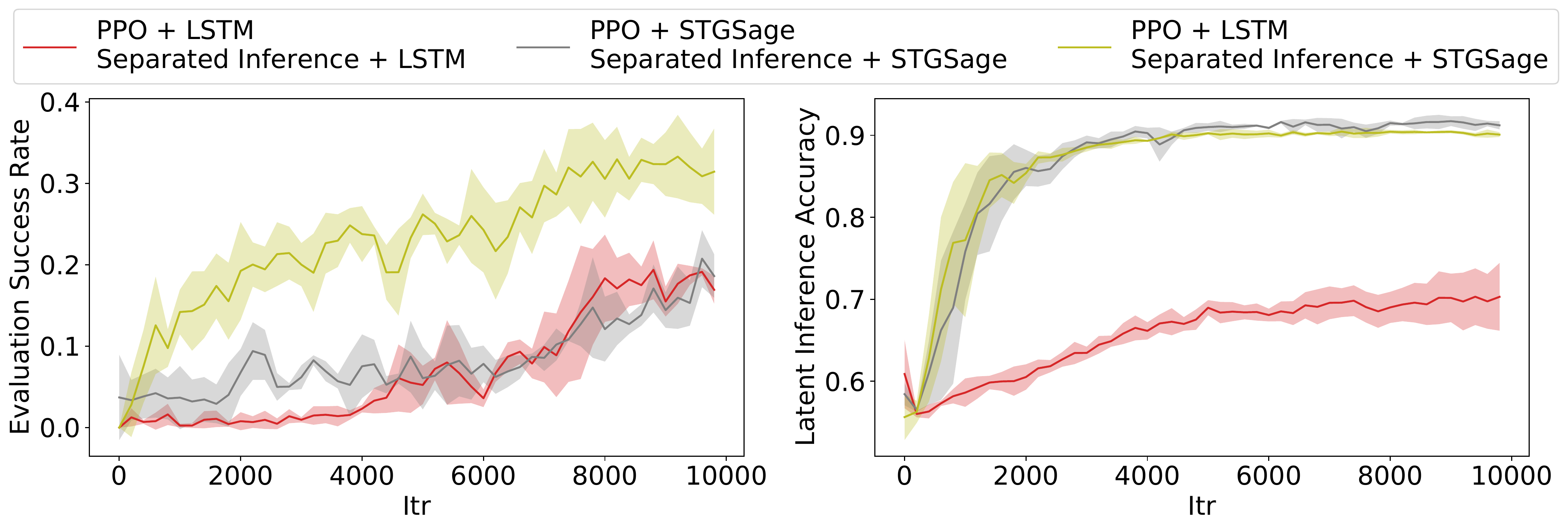}
	\caption{\small \textbf{Network comparison}: Influence of the network structure on reinforcement learning and intention inference with separated inference architecture. Left: Test time success rate. The success rate using a mixed network structure (yellow line) is 70\% higher than baselines. Right: Latent state inference accuracy. The inference accuracy using STGSage is 40\% higher than LSTM.
    \normalsize}\label{fig:network_compare}
    \vspace{-0.2cm}
\end{figure}

\begin{figure}[!tbp]
	\centering
	\includegraphics[width=0.8\columnwidth]{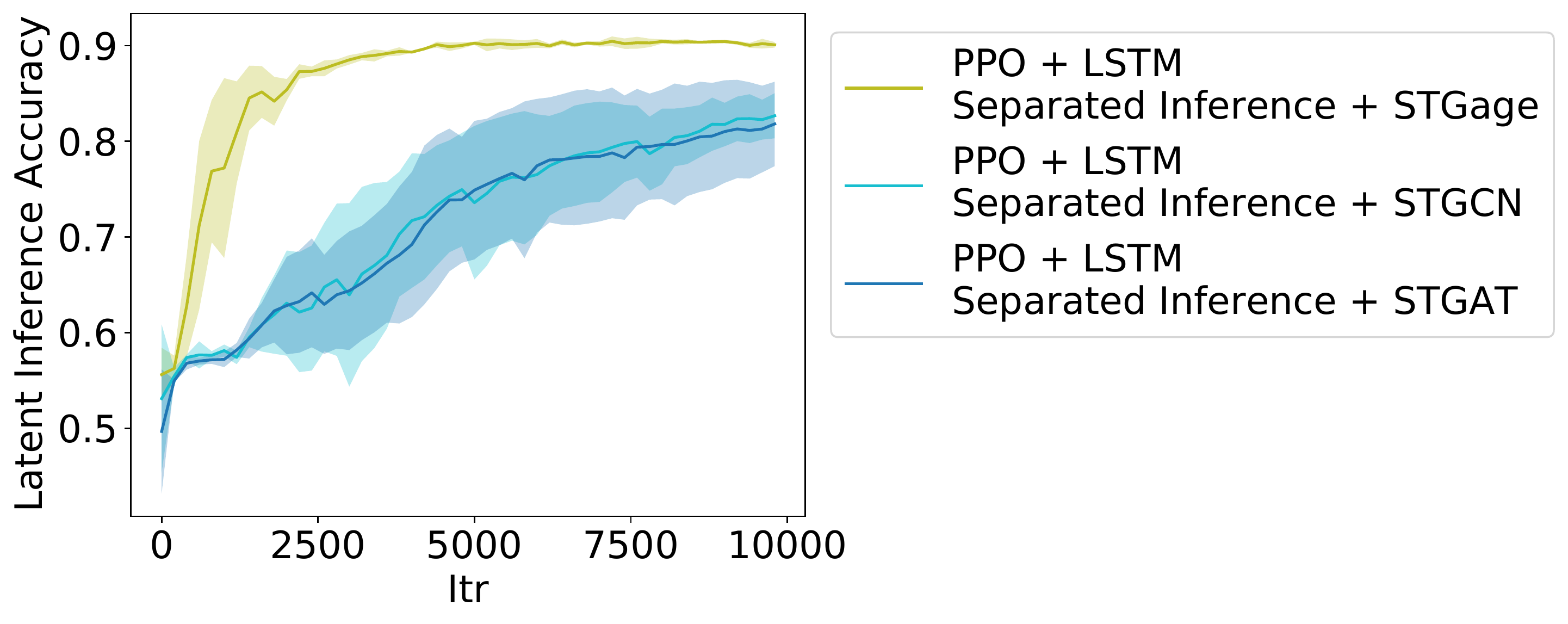}
	\caption{\small \textbf{GNN comparison}: Influence of the GNN structure on the latent state inference with separated inference architecture. The final inference accuracy of STGSage is 10\% higher than the best baseline. \normalsize
    }\label{fig:gnn_compare}
\vspace{-0.5cm}
\end{figure}

\subsection{Influence Passing Modeling}
\label{subsec: network_compare}
As the separated inference structure gives the best performance with both LSTM and STGSage networks, we further compare the influence of explicitly modeling the influence passing by STGSage under the separated structure. 
The results are shown in \cref{fig:network_compare}.
Comparing the results of using LSTM or STGSage for both tasks (the red and grey lines in \cref{fig:network_compare}), we see that the two networks converge to a similar success rate of 0.2. The inference accuracy of using STGSage is 40\% higher than using the LSTM.

Since the success rate is also influenced by the inference accuracy, we test an additional structure where we use LSTM as the policy network and STGSage as the latent inference network (the yellow lines in \cref{fig:network_compare}). 
The mixed network structure shows a significant improvement on the reinforcement learning performance with a similar inference accuracy as STGSage. This indicates that the LSTM works better than STGSage as a policy network when the latent state of other vehicles are known.
The opposite comparison results of the two network structures on vehicle control and latent inference is likely due to the different nature of the two tasks.
The latent inference is more decentralized where the network needs to focus on the local interactions of each surrounding vehicle with its neighbors. Such local interactions have a shared structure which are better captured by the graph representation and convolutions.
On the contrary, the reinforcement learning task is more centralized on the ego vehicle. In this case, the LSTM structure provides more flexibility on focusing on the ego-relevant features.
With the modularized design of our framework, we could easily combine the best of two different network structures.
The final success rate achieved of 30\% does not mean that there is a collision in 70\% of the test cases. In most unsuccessful cases, the ego vehicle waits at the intersection because a collision-free right-turn is impossible due to the dense traffic.

\subsection{Interpretability}
\begin{figure}[!tbp]
	\centering
	\begin{subfigure}[t]{0.45\columnwidth}
		\centering
		\includegraphics[width=\columnwidth]{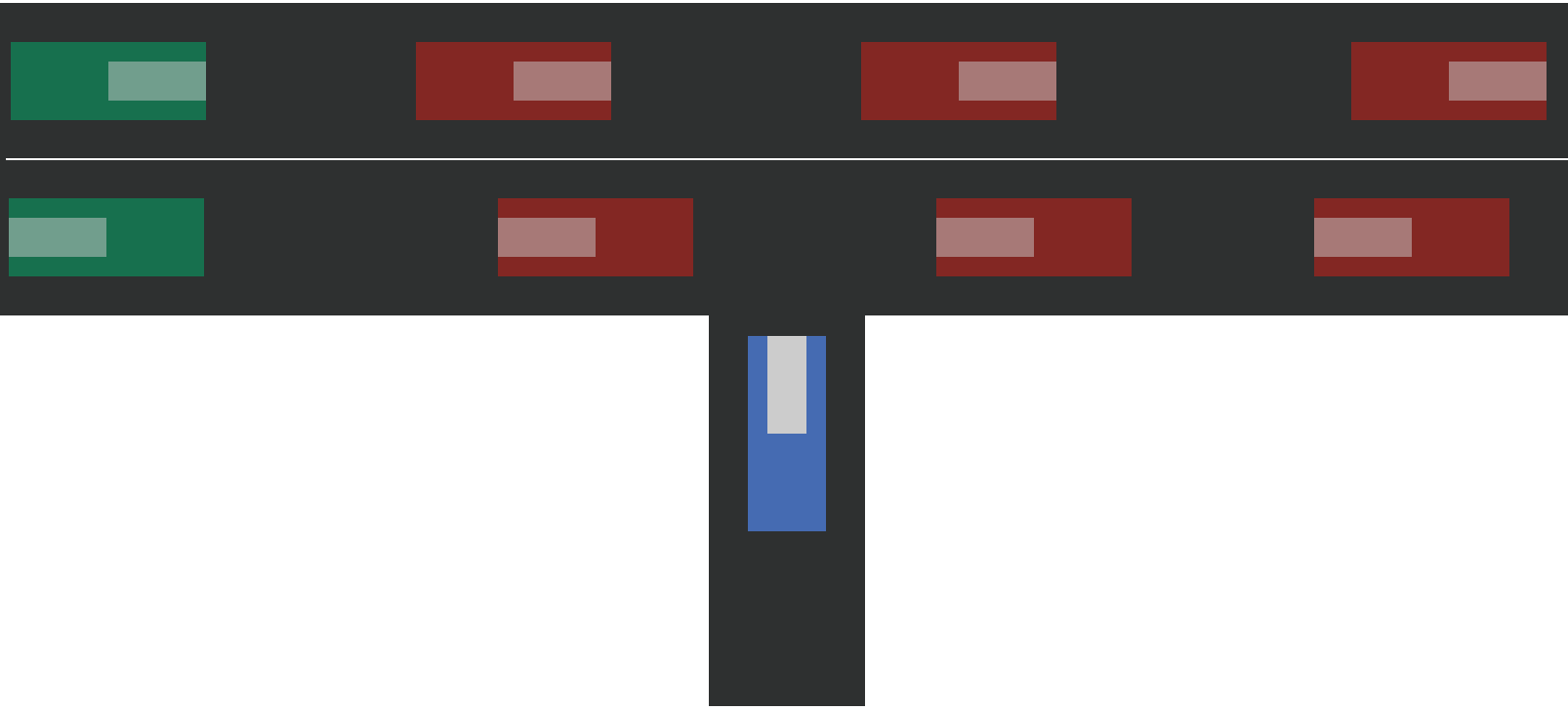}
		\caption{\small PPO + LSTM \normalsize}\label{fig:ppo_lstm_rollout}		
	\end{subfigure}
	\begin{subfigure}[t]{0.45\columnwidth}
		\centering
		\includegraphics[width=\columnwidth]{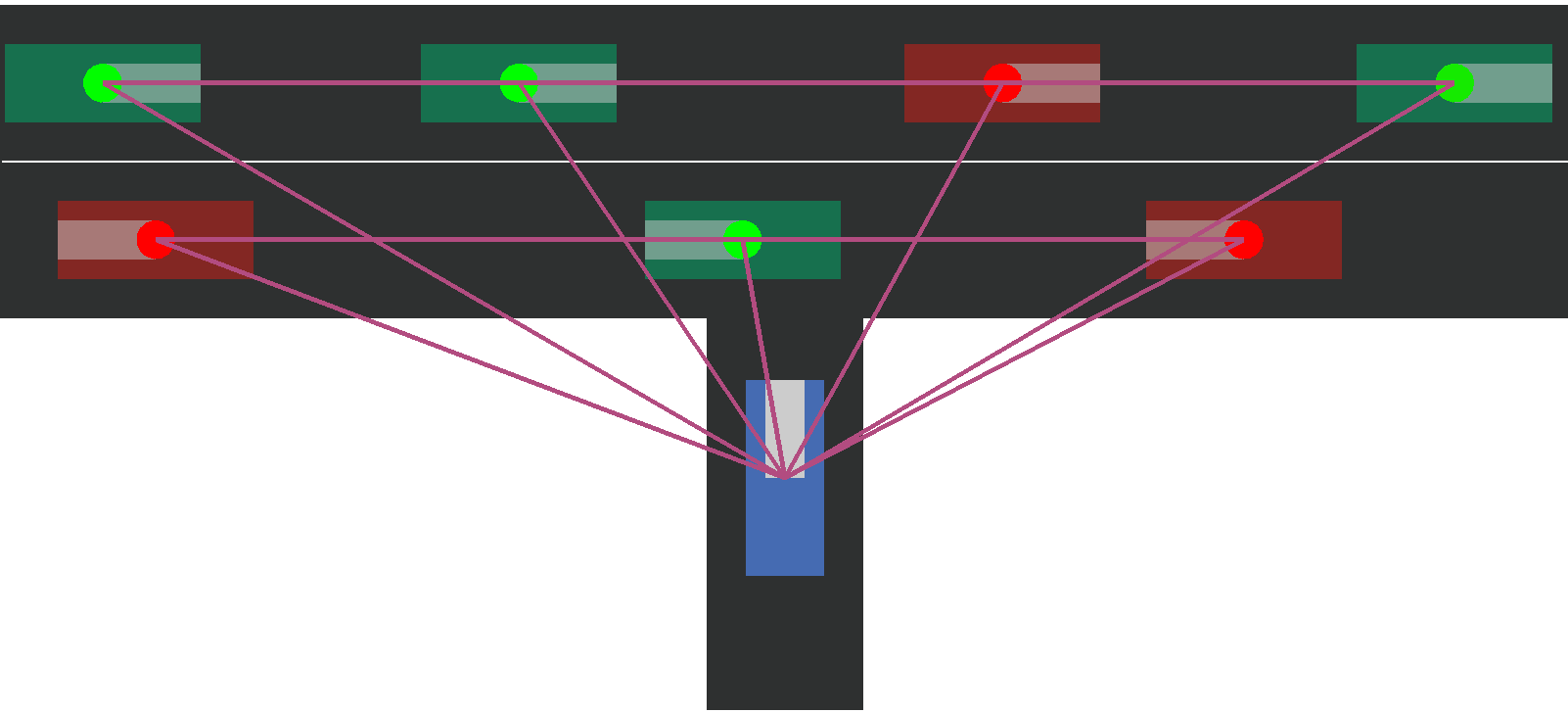}
		\caption{\small PPO + LSTM + Separated Inference +STGSage \normalsize}\label{fig:supsep2_lstmgnn_rollout}
	\end{subfigure}
    \vspace{-0.2cm}
	\caption{\small \textbf{Interpretability}: The latent inference and graph representation add additional interpretability to the learned policy. The color of the vehicle indicates the hidden latent mode (green: conservative; red: aggressive). The color of the circle at the center of each vehicle indicates the inferred latent mode. The purple lines connecting different vehicles represent the message passing paths.\normalsize}\label{fig:inter}
 	\vspace{-0.6cm}
\end{figure}

We compare the interpretability of different structures by examining how much interpretable information we could get from the trained policy. As shown in \cref{fig:inter}, the vanilla PPO is close to a black box where we could only get the action distributions given observations. The auxiliary latent inference gives us information about how the policy infers the latent states of surrounding vehicles. The graph representation used by the STGSage also gives additional interpretability on the influence passing structure inside the network.

\subsection{Ablation Study: Graph Neural Network Structure}
\label{subsec:gnn_compare}
We compare the performance of different graph convolutional layers used in processing the spatial-temporal graph. Under the separated inference structure, we fix the policy network as LSTM and vary the convolutional layers between GraphSAGE \cite{graphsage} (STGSage), GAT \cite{gat} (STGAT), and GCN \cite{kipf2016semi} (STGCN) in the latent inference network. The results are shown in \cref{fig:gnn_compare}. The latent inference accuracy from the STGSage is significantly higher than that of STGCN and STGAT. By concatenating the self node embedding with the aggregated neighbour embedding, the GraphSage convolution is able to flexibly distinguish the self node embedding from its neighbors in the embedding update. Thus, it better models the egocentric decision process of the surrounding driver.

\subsection{Robustness to Distribution Shift}
\begin{figure}[!tbp]
	\centering
	\includegraphics[width=\columnwidth]{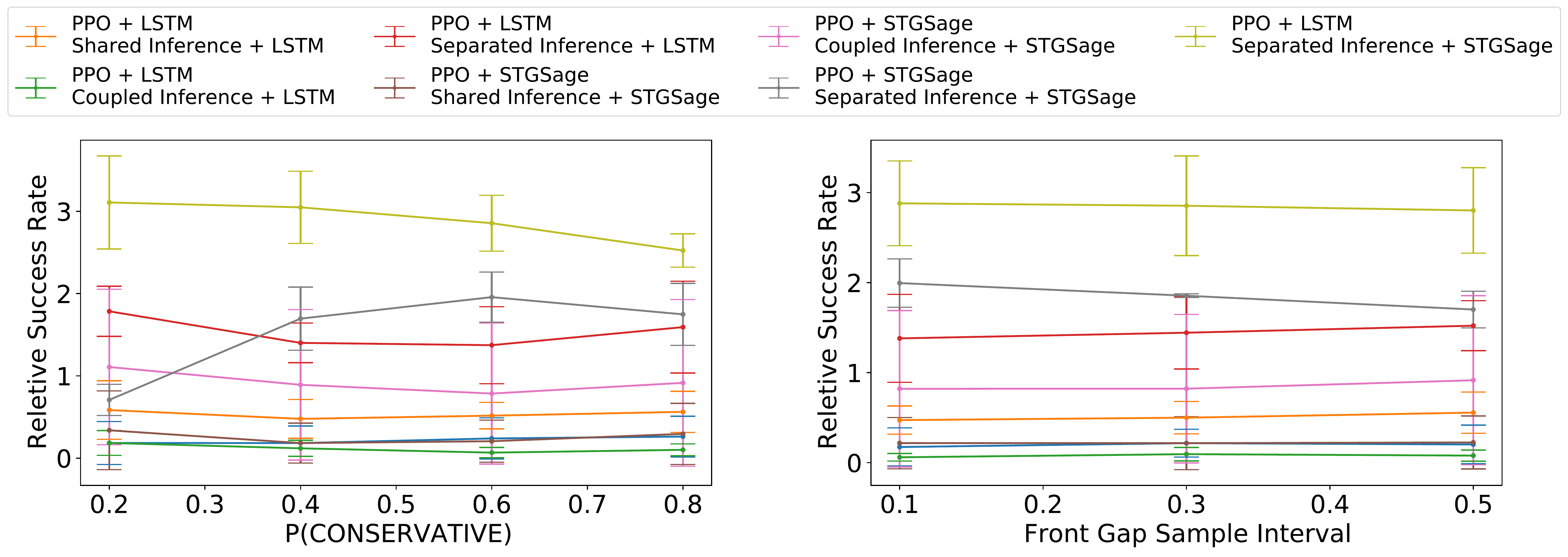}
	\caption{\small \textbf{Robustness against distribution shift}. Left: Performance change against the latent state distribution. 
    $P(\textsc{conservative})=0.5$ at training time. 
	Right: Performance change against the surrounding driver model; 
	The interval length is 0.3 at training time.
	The test time success rate is reported relative to the average success rate over all approaches at the same setting. 
	The proposed structure (yellow) shows the best and most robust performance.
    \normalsize}\label{fig:drift}
    \vspace{-0.6cm}
\end{figure}

We further test the robustness of the different approaches against the underlying distribution change in the environment at the test time. 
We first vary the latent distribution, $P(\textsc{conservative})$, which is the probability that a surrounding driver being conservative. $P(\textsc{conservative})$ is fixed at 0.5 at training time. As $P(\textsc{conservative})$ becomes smaller, more drivers in the main lane are likely to be aggressive, and the difficulty for making a successful right turn increases. As shown in the left part of \cref{fig:drift} (a), as the environment becomes harder, the advantage of the proposed approach (yellow line) becomes larger. 

We also investigate the influence of changing the surrounding driver model by varying the interval length of the uniform distribution where the surrounding driver samples its desired front gap. The mean of the distribution is kept the same. As shown in the right part of \cref{fig:drift}, the proposed approach (yellow line) has the best as well as the most robust performance against the change of the distribution.

\section{Conclusions}
We studied the effect of latent state inference as an auxiliary task on the performance of the reinforcement learning in navigating through an uncontrolled T-intersection. 
We propose a novel framework which combines the latent inference by supervised learning and the vehicle control by reinforcement learning with minimal structural coupling.
The proposed framework improves the navigation success rate by at least 70\%, and the latent inference accuracy by 10\%.
We also investigated the application of the graph representation and graphical neural networks (GNNs) on processing traffic information. We propose STGSage, a spatial-temporal GNN structure, which efficiently learn the spatial interactions as well as the temporal progressions of different traffic participants.
STGSage improves the latent inference accuracy by 40\% in the T-intersection scenario.
While the success of the deployment of the learned policy to real vehicles relies on accurate simulation models during training, we expect the proposed framework to better generalize with high-level latent state and action spaces.

\printbibliography

\addtolength{\textheight}{-12cm}   





\end{document}